\documentclass[logo, address]{trfr}

\usepackage{microtype}
\usepackage{graphicx}
\usepackage{subcaption}
\usepackage{booktabs} 
\usepackage{pgfplots}
\usepackage{tikz}
\usepackage{svg}
\usepackage{hyperref}

\usepackage{amsmath}
\usepackage{amssymb}
\usepackage{mathtools}
\usepackage{amsthm}
\usepackage{tabularx}
\usepackage{booktabs}
\usepackage{colortbl}
\usepackage{xcolor}
\usepackage{array}
\usepackage{caption}
\usepackage{makecell}
\usepackage{rotating}
\usepackage{multirow}
\usepackage{graphicx}

\usepackage[most]{tcolorbox}
\usepackage{enumitem}
\usepackage{xcolor}

\renewcommand{\arraystretch}{1.3}

\usepackage{tikz}
\usepackage[T1]{fontenc}
\usepackage{cancel}
\usepackage{tcolorbox}
\usepackage{varwidth}
\tcbuselibrary{skins, breakable}
\usetikzlibrary{arrows.meta}
 
\definecolor{defined}{HTML}{F5A623}
\definecolor{crossref}{HTML}{E8A0A0}
\definecolor{capcolor}{HTML}{7ECEC4}
\definecolor{boxbg}{HTML}{F5F2EC}
\definecolor{issuebg}{HTML}{EFEFEA}
\definecolor{rulecol}{HTML}{CCCCCC}
 
\newcommand{\hl}[2]{\colorbox{#1}{\strut\sffamily #2}}
\newcommand{\hlst}[2]{\colorbox{#1}{\strut\sffamily\textcolor{black}{\cancel{#2}}}}
 
\newcommand{\issuebadge}[2]{%
  \begin{tcolorbox}[
    colback=#1, colframe=#1, arc=3pt,
    boxsep=1pt, left=4pt, right=4pt, top=2pt, bottom=2pt,
    width=\linewidth,
  ]
    \small\bfseries\sffamily\color{white}#2
  \end{tcolorbox}%
}

\usepackage{enumitem}
\usetikzlibrary{positioning, calc, shapes.geometric}

\definecolor{c1}  {RGB}{  0, 110,  80}   
\definecolor{c1bg}{RGB}{230, 247, 240}
\definecolor{c2}  {RGB}{ 75,  60, 175}   
\definecolor{c2bg}{RGB}{238, 235, 255}
\definecolor{c3}  {RGB}{180,  50,  40}   
\definecolor{c3bg}{RGB}{253, 235, 232}
\definecolor{gold}{RGB}{175, 130,  10}
\definecolor{goldbg}{RGB}{255, 248, 220}
\definecolor{neutral}{RGB}{100, 95, 90}
\definecolor{neutralbg}{RGB}{240, 237, 232}
\definecolor{rulecolor}{RGB}{200, 195, 188}

\tcbset{
  stagecard/.style={
      enhanced, sharp corners,
      boxrule=0pt, frame hidden,
      borderline west={3pt}{0pt}{#1},
      colback=#1bg,
      left=10pt, right=10pt, top=9pt, bottom=9pt,
      width=4.8cm,
      equal height group=stages,
      height=4.5cm,
    },
  stagecard/.default=c1,
}

\tcbset{
  outtag/.style={
    enhanced, sharp corners,
    boxrule=0.7pt, colframe=#1, colback=#1bg,
    left=4pt, right=4pt, top=1pt, bottom=1pt,
    fontupper=\scriptsize\bfseries\color{#1},
    width=3.8cm, halign=center,
    nobeforeafter,
  },
  outtag/.default=c1,
}

\usepackage[
  backend=biber,
  natbib=true,
  sorting=none,
  style=numeric,
  maxnames=5,
  minnames=1
]{biblatex}
\usepackage{wrapfig}

\usepackage{tocloft}
\usepackage{placeins}
\usepackage{longtable}

\definecolor{h0}{RGB}{239, 83, 80}    
\definecolor{h1}{RGB}{239,130,110}    
\definecolor{h2}{RGB}{239,170,140}    
\definecolor{h3}{RGB}{239,154,154}    
\definecolor{h4}{RGB}{255,224,130}    
\definecolor{h5}{RGB}{220,237,200}    
\definecolor{h6}{RGB}{165,214,167}    
\definecolor{h7}{RGB}{102,187,106}    
\definecolor{h8}{RGB}{67,160, 71}     
\definecolor{h9}{RGB}{46,125, 50}     

\newcommand{\heatcell}[1]{%
  \ifdim #1 pt < 0.1pt \cellcolor{h0}%
  \else\ifdim #1 pt < 0.2pt \cellcolor{h1}%
  \else\ifdim #1 pt < 0.3pt \cellcolor{h2}%
  \else\ifdim #1 pt < 0.4pt \cellcolor{h3}%
  \else\ifdim #1 pt < 0.5pt \cellcolor{h4}%
  \else\ifdim #1 pt < 0.6pt \cellcolor{h5}%
  \else\ifdim #1 pt < 0.7pt \cellcolor{h6}%
  \else\ifdim #1 pt < 0.8pt \cellcolor{h7}%
  \else\ifdim #1 pt < 0.9pt \cellcolor{h8}%
  \else \cellcolor{h9}%
  \fi\fi\fi\fi\fi\fi\fi\fi\fi
  #1%
}
\theoremstyle{plain}

\theoremstyle{definition}

\theoremstyle{remark}

\newcommand{\ourbenchmark}{\textsc{ContractScrub}}

\usepackage[textsize=tiny]{todonotes}

\shorttitle{ContractScrub}

\title{\ourbenchmark: A benchmark for final review of legal contracts}

\abstract{

Legal work, with its heavy reliance on processing large amounts of text, is often considered one of the domains most exposed to the use of LLMs.  Contract ``scrubbing,'' the final review of transactional agreements for errors and inconsistencies, is a particularly suitable task for automation, because it is routine, painstaking work requiring detailed attention to long documents. Scrubbing also seems to align naturally with the general capabilities expected of frontier LLMs around long-context reasoning, consistency checking, and named entity recognition (NER). Despite the economic value and potential for automation, no formal evaluations of LLMs performing contract scrubbing have been conducted. We introduce \ourbenchmark, the first benchmark designed to evaluate contract scrubbing capabilities, comprising contracts hand-crafted by experienced lawyers over diverse error categories such as misuse of defined terms, incorrect references, and inconsistent language. Frontier models perform surprisingly poorly with only one model reaching 0.75 macro average recall despite strong performance on seemingly related general benchmarks, demonstrating the practical limits of current models and the importance of narrowly targeted, domain-specific benchmarks for measuring real-world impact.

\begin{tabular}{@{}c l@{}}
{\textbf{Dataset:} \href{https://huggingface.co/tri-fair-lab/contract_scrub}{\texttt{https://huggingface.co/tri-fair-lab/contract\_scrub}}}\\
\end{tabular}

}

\author[1]{Yejin Bang$^{*}$}
\author[1]{Kirsty Fielding$^{*}$}
\author[1]{Brandan Oliver$^{*}$}
\author[1]{Brian Birke$^{*}$}
\author[1,2]{Nabeel Seedat$^{*}$}
\author[1,2]{Andrew M. Bean$^{*}$}
  
\affiliation[1]{Thomson Reuters Foundational Research, London, UK}
\affiliation[2]{Imperial College London, UK}

\correspondence{\{first.last\}@thomsonreuters.com}
\contribution[*]{Equal contribution}

\begin{document}

\maketitle

\section{Introduction}

\begin{wrapfigure}{r}{0.5\linewidth}
  \scalebox{0.58}{%
    \begin{tikzpicture}[font=\sffamily]
\node[anchor=north west] (doc) at (0,0) {%
  \begin{tcolorbox}[
      colback=boxbg, colframe=rulecol,
      arc=6pt, boxrule=0.6pt,
      width=6.8cm,
      top=8pt, bottom=8pt, left=10pt, right=10pt,
  ]
    \begin{center}
      \textbf{SERVICE AGREEMENT}\\[2pt]
      {\scriptsize January 1, 2024}
    \end{center}
    \tcbline
    \textbf{1. DEFINITIONS}\\[4pt]
    ``Agreement'' means this current agreement, inclusive of Schedules A-C.\vspace{0.25em}\\
    ``Services'' means consulting services as listed in \hlst{crossref!35}{Schedule D}.
    \tcbline
    \textbf{2. OBLIGATIONS}\\[4pt]
    2.1 The Supplier shall provide Services to the Client at the prices specified in the Schedules contained in the \hl{defined}{agreement.}
    \tcbline
    \textbf{3. TERMINATION}\\[4pt]
    Either party may terminate the Agreement if the other commits a \hl{capcolor!60}{Material Breach}.
    \vspace{6pt}
    \begin{center}{\scriptsize --- 1 ---}\end{center}
  \end{tcolorbox}%
};
\node[anchor=west] (issues) at ([xshift=0.8cm]doc.east) {%
  \begin{tcolorbox}[
      colback=issuebg, colframe=rulecol,
      arc=6pt, boxrule=0.6pt,
      width=6.2cm,
      top=8pt, bottom=8pt, left=10pt, right=10pt,
  ]
    \begin{center}\textbf{Identified issues}\end{center}
    \issuebadge{defined}{Not using defined terms}
    \vspace{2pt}
    Using ``agreement'' in \S2.1 introduces questions about what agreement is at issue 
    \vspace{6pt}
    \issuebadge{crossref}{Incorrect reference}
    \vspace{2pt}
    ``Schedule D'' in \S1 is not part of the Agreement
    \vspace{6pt}
    \issuebadge{capcolor}{Missing definition}
    \vspace{2pt}
    ``Material Breach'' referenced in \S3 is capitalized, implying it is a defined term, but it is not defined
  \end{tcolorbox}%
};
\draw[-{Stealth[length=5pt,width=4pt]}, gray!60, line width=1pt]
    (doc.east) -- (issues.west);
\end{tikzpicture}
  }
\caption{An example of errors that should be found in the scrubbing process.}
\end{wrapfigure}
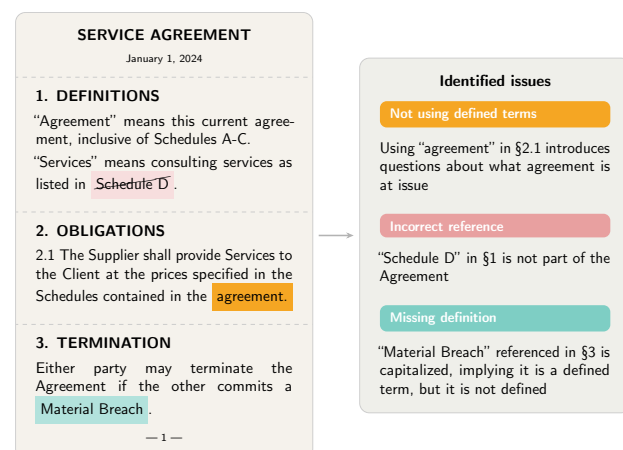

The practice of law ranks highly among the professions where AI is expected to have the most potential for economic impact \cite{laneImpactArtificialIntelligence,felten2023languagemodelerslikechatgpt}. Benchmarks such as LEXam \cite{fan2025lexam} and Stanford Legal Bench \cite{guha2023legalbench} are typically used to assess progress in legal capabilities like reasoning. However, translation between capability evaluations and real-world impact can often be limited \cite{schwartz2025reality, bean2025measuringmattersconstructvalidity}. By directly testing models on economically valuable tasks, narrowly targeted benchmarks with high ecological validity offer a more reliable measure of the potential of current legal AI systems. 

Transactional law practices involve the negotiation, drafting, and execution of legally binding agreements and navigation of the complex relationships of the parties. Contracts, as a subset of transactional law, place heavy demands on precision and nuance. Small details can affect legal interpretation, a famous example being the lack of an Oxford comma in the statutes governing Maine employment contracts, which led to a multi-million dollar settlement \footnote{See O'Connor v. Oakhurst Dairy, 851 F.3d 69 (1st Cir. 2017); see also https://www.bbc.co.uk/worklife/article/20180723-the-commas-that-cost-companies-millions}. Contract scrubbing occupies a peculiar position in transactional law practices, widely recognized as essential but also generally tedious to carry out. At the very end of (and often throughout) a deal, when negotiations are complete but before signatures are exchanged, lawyers and paralegals will ``scrub'' a contract; making a final pass to remove any outstanding errors or inconsistencies. This exercise offers high ecological validity as it is a discrete, well-scoped task with a clear ground truth directly replicating work performed by professionals under realistic conditions.

Though critically necessary, scrubbing is meticulous, repetitive work often performed under intense pressure from the demands of closing a deal, making this routine task error-prone in practice. At the same time, many elements of contract scrubbing (e.g. identifying defined terms, checking consistency of usage, and verifying section references) seem to fit naturally within the capabilities of LLMs around long-context reasoning, consistency checking, and named entity recognition (NER). Legal practitioners frequently approach this risk-control function as essential but tedious work. If the scrub could be automated reliably, practitioners could redirect their attention to other matters: identifying issues that demand legal judgment, advising clients on risk, and advancing negotiations. Attorneys could also perform automated scrubbing more frequently throughout the negotiation of a contract, reducing the risk of last-minute changes disrupting an otherwise settled deal.

The clear potential for AI-assisted contract scrubbing makes it a natural domain for a targeted, task-specific benchmark, but, to the best of our knowledge, no such benchmark exists. Current contract-related benchmarks focus largely on reasoning about contracts rather than reviewing them for precision and consistency. CUAD \cite{hendrycks1cuad} tests whether models can identify legally significant provisions such as governing law, exclusivity, and non-compete clauses. \citeauthor{liu-etal-2025-contracteval} introduced ContractEval, a benchmark focused on evaluating LLMs' ability to identify clause-level legal risks in commercial contracts. And, general-purpose tasks in NLP like "needle-in-a-haystack" \cite{kamradt2023needleinhaystack}, are more similar to the actual work required for scrubbing, but do not share the idiosyncrasies of the legal domain.

\begin{table*}[]
\centering
\caption{Scrub Review Categories. Each tested contract contains different categories of errors. }
\renewcommand{\arraystretch}{1.4}
\small
\begin{tabular}{p{5cm}|p{11cm}}
\toprule
\textbf{Category} & \textbf{Explanation} \\
\midrule
Defined Terms & Words or phrases formally defined in the agreement with a designated meaning, often capitalized, underlined, bolded, italicized, or in quotations. \\
Undefined Capitalized Terms & Terms that are capitalized or otherwise treated as a defined term in an agreement but not formally defined. \\
Uncapitalized Defined Terms & Formally defined terms appearing in lowercase when they should be capitalized.  \\
Incorrectly Capitalized Terms In Context & Terms that have a definition but are capitalized in a context where they are not being used as a defined term.\\
Unused Defined Terms & Defined terms that are not used outside of the specific instance where they are defined.\\
Terms Defined Multiple Times & Terms that are defined more than one time in an agreement. \\
Incorrect Section, Article, or Paragraph References & Internal cross-references that assign an incorrect section, article, or paragraph of the agreement. \\
Incorrect Party References & Instances where a party is referred to by the wrong party name or role.  \\
Inconsistent Language & Language in the agreement that directly contradicts itself or other language elsewhere in the agreement. \\
\bottomrule
\end{tabular}
\label{tab:scrub_categories}
\end{table*}

To address this gap, we introduce \ourbenchmark, the first benchmark designed to evaluate LLM performance on the various (including final) review stages of the contract or deal lifecycle. ContractScrub comprises 3,014 annotated tasks across 44 contracts drawn from CUAD \cite{hendrycks1cuad}, including scrubbing elements and errors, e.g., defined term inconsistencies, capitalization errors, and cross-reference failures. The contracts are hand annotated by experienced lawyers to include representative test issues. 

We evaluate 9 frontier and open-weight models of varying families and sizes on \ourbenchmark~and find that contract scrubbing remains a substantially harder task for LLMs than its individual components might suggest. The best-performing model, GPT-5.5, reaches a macro-average recall score of only 0.750, and all F1 scores are below 0.650. Performance is uneven across issue types: models handle categories with explicit lexical signals (e.g., defined term; $\mu=$ .835) much better than those requiring inference of intent within context (e.g., Incorrect Capitalization in Context; $\mu=$.427). Beyond the practical value of filling the measurement gap, our benchmark offers a valuable theoretical insight. Despite many of the tasks required for scrubbing sharing a strong similarity with conventional NLP tasks, most frontier models fall short of the performance that would be expected based on their general capabilities, highlighting the value of domain-specific benchmarking as a practice. Furthermore, enabling reasoning yields only moderate gains, concentrated in categories that require full-document term consistency rather than deeper legal interpretation. These findings underscore the need for targeted benchmarks in professional domains, and we release \ourbenchmark~to contribute to that effort.

\section{Related Work}

\paragraph{Legal Benchmarks for LLMs}
Evaluating the legal capabilities of LLMs is an active area of research, with recent benchmarks aimed at measuring broad legal competence across diverse tasks. LEXam~\cite{fan2025lexam}, Stanford LegalBench~\cite{guha2023legalbench}, LawBench~\cite{fei2024lawbench}, and LexGLUE~\cite{chalkidis2022lexglue} exemplify this general-purpose approach, evaluating LLMs in collections of legal problems that test skills such as legal knowledge, reasoning, and interpretation. These suites are valuable for measuring overall progress in legal AI, but their tasks typically probe narrow skills in isolation rather than end-to-end review of a single long document. Contract scrubbing is one such workflow: it requires sustained attention to long, highly structured documents, where small textual inconsistencies can have legal or commercial consequences, or both. 

A closer line of work does evaluate LLMs in contract-specific tasks. CUAD~\cite{hendrycks1cuad} tests whether models can identify important clauses and legal attributes in thousands of annotated examples of commercial contracts, while MAUD~\cite{wang-etal-2023-maud} provides a similarly large-scale benchmark for locating deal-points in merger agreements. Both cast contract review largely as reading comprehension over predefined legal categories, requiring models to locate relevant provisions or determine whether particular deal points are present --- e.g ``is there a non-compete clause?'' ContractNLI~\cite{koreeda2021contractnli} evaluates contractual reasoning asking whether contract provisions entail, contradict, or leave undetermined a set of hypotheses. The Lease benchmark of \citeauthor{leivaditi2020benchmark} evaluates NER and red-flag detection on residential lease agreements, and ContractEval~\cite{liu-etal-2025-contracteval} extends this to legal-risk identification across a wider set of commercial agreements. Both move closer to practical review, but still assess classification of provisions against a fixed taxonomy of risk types. In contrast, \ourbenchmark{} evaluates defect identification as a recall-sensitive, full-document task.
Rather than answering a supplied question, classifying a provision against a known taxonomy, or evaluating a predefined hypothesis, the model is expected to surface drafting, consistency, and document-hygiene defects in a contract, which, if missed, can have material legal and practical consequences. In short, prior contract benchmarks largely evaluate ``is X here?'' under a known schema. \ourbenchmark{} evaluates over the whole document ``what is wrong with this relative to the contract’s own internal conventions?''

\paragraph{Related Non-Legal Tasks}

Contract scrubbing requires the joint application of several general-purpose LLM capabilities, including long-context reasoning, precise localization, consistency checking, and referential understanding, in a structured legal setting. Existing non-legal benchmarks probe related capabilities, but generally in more isolated settings. Long-context benchmarks \cite{hsieh2024ruler,ling2025longreason} evaluate whether models can reason over extended inputs, a prerequisite for contract scrubbing because relevant errors may be sparsely distributed across a lengthy document. Needle-in-a-Haystack-style evaluations \cite{kamradt2023needleinhaystack, chen-etal-2025-longleader} require models to find a single embedded fact within a long context, which parallels the localization demands of scrubbing but not the requirement to surface \emph{all} relevant defects based on the broader context of the document. FaithEval \cite{ming2025faitheval} includes inconsistency detection over short passages, resembling conflicting-definition or inconsistent-term errors. However, it treats consistency primarily as a binary classification problem, rather than requiring models to find, locate, and contextualize inconsistencies across a long document. IdentifyMe \cite{manikantan-etal-2025-identifyme} benchmarks entity reference resolution, a capability relevant to detecting incorrect party names and section references, but operates on general-domain text. Each of these benchmarks isolates an individual component capability. \ourbenchmark{}  is designed to evaluate their collective application to comprehensive defect identification and localization in full-length contracts.

\section{\ourbenchmark~Benchmark}

Contract review involves reading and understanding a contract thoroughly to identify errors, analyse risks, and ensure consistency. Throughout the negotiation process and certainly at the final stage of review, attorneys conduct a dedicated pass to eliminate residual errors and inconsistencies -- a process commonly known as ``scrubbing.'' Seemingly small errors in contracts are time-intensive to identify and review and can be highly consequential, requiring legal practitioners to catch them in the contract-drafting lifecycle. However, they remain difficult to identify under the time pressures typically accompanying such a routine task (that is, nearing deal closing or contract execution) and because manual review is repetitive, subject to fatigue, and complicated by multiple negotiated drafts, schedules, and amendments. Automating or augmenting legal practitioners at this stage with LLMs offers a practical path to freeing practitioners for higher-value work by reducing avoidable drafting mistakes, improving first-pass review, supporting junior attorneys, accelerating quality control, and allowing senior lawyers to spend more time on issues requiring their judgment and experience, all the while upholding the high standards commensurate with the legal profession.

\subsection{Task Overview}

\begin{figure*}
    \centering
    \caption{ContractScrub construction pipeline. Contracts are sourced and screened for existing structural issues (Stage 1),  annotated for existing drafting errors (Stage 2), and augmented with targeted additional errors (Stage 3) to produce the gold answer. All steps are carried out by experienced lawyers.}
    \label{fig:pipeline}
    \scalebox{1.0}{%
    \includegraphics[width=0.7\linewidth]{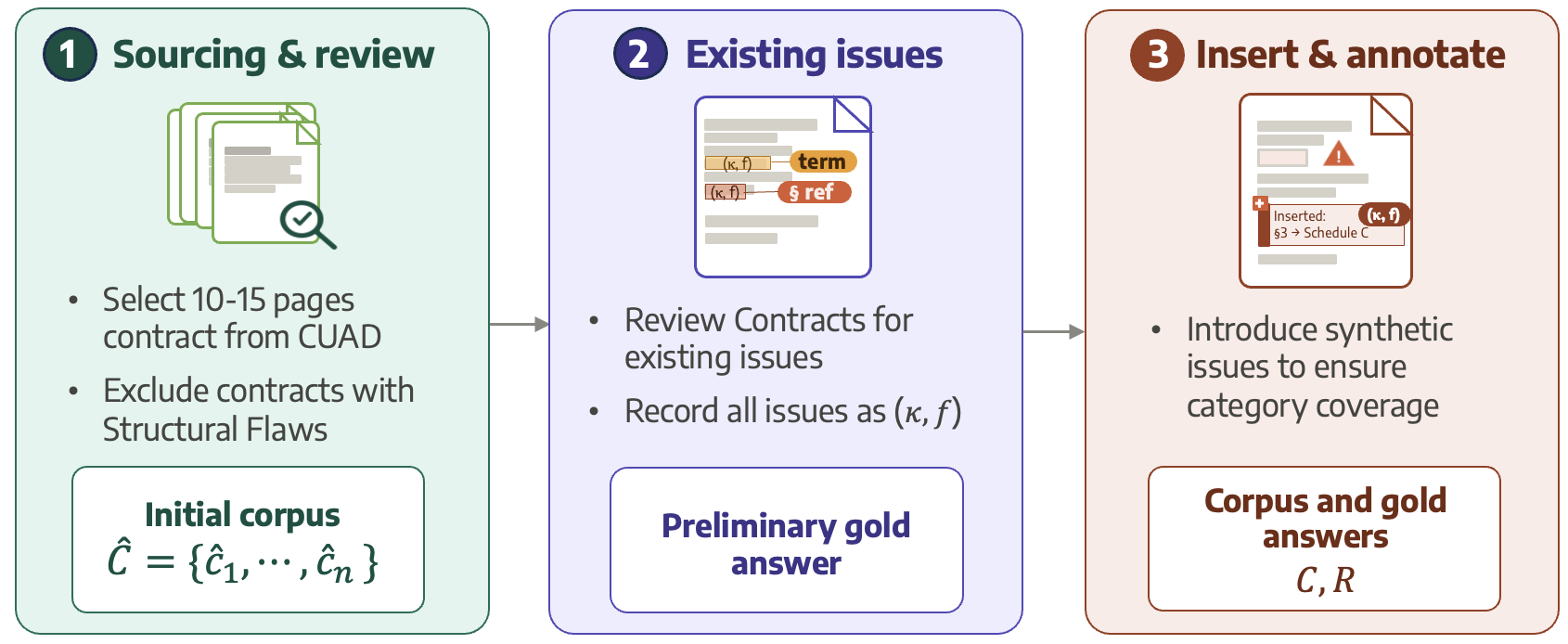}
  }
\end{figure*}

\ourbenchmark~is the first benchmark evaluating LLM performance on the scrubbing pass. The task is defined as follows. Let $\mathcal{C} = \{c_1, \dots, c_n\}$ be a corpus of $n$ contracts.  For each contract $c_i$, a gold annotation $R_i$ is a multiset of tuples, where each tuple $\mathbf{r} = (\kappa, \mathbf{f})$ consists of a category label $\kappa \in \mathcal{K}$ and a category-specific field vector $\mathbf{f}$. The category set $\mathcal{K}$ comprises nine elements covering one defined-term extraction category and eight drafting-error categories (Table \ref{tab:scrub_categories}). The field vector $\mathbf{f}$ encodes the category-specific fields for each instance --- for example, $\mathbf{f} = (\textit{term}, \textit{location})$ for most term-level categories, and $\mathbf{f} = (\textit{location}_1, \textit{location}_2)$ for relational categories such as inconsistent terms. 
The task is designed to simulate the scrubbing pass as performed by actual attorneys in practice: given contract $c_i$ and an instruction prompt $I$ as input, a model must produce a predicted multiset $\hat{R}_i$ of tuples drawn from the same schema. Formally, each model $\mathcal{M}$ produces a
predicted annotation:
\[
    \hat{R}_i = \mathcal{M}(I, c_i),
\]
Performance is measured by comparing $\hat{R}_i$ against the gold $R_i$ across all categories and contracts.

\subsection{Dataset}
To systematically evaluate scrubbing capability of LLMs, we construct dedicated annotated data (Figure \ref{fig:pipeline}). The schema of categories is designed by licensed attorneys with professional experience practicing and litigating contract law. Then, the further annotation and creation of the dataset is conducted by 9 different lawyers with experience in various forms of commercial, corporate, and contract law. All the lawyers have been practicing law for at least 8 years, with 8 having more than 10 years experience and 6 having more than 15 years.

\ourbenchmark~consists of 3,014 annotated tasks across 9 categories drawn from 44 contracts. Each contract $c_i$ has corresponding gold annotation $R_i$, which consists of different categories explained in Table~\ref{tab:scrub_categories}.
The guiding principles in creating this benchmark are improving contract hygiene (precision and consistency), reducing the risk of misinterpretation and ambiguity, and preserving client confidence by uncovering drafting defects that can have outsized impacts on contractual meaning, clarity, and outcomes.

\paragraph{Scrub Categories}

Each gold answer covers nine annotation categories listed in Table \ref{tab:scrub_categories}: one defined-term extraction category and eight drafting-error categories. The categories were selected because they are common, concrete, and potentially impactful on a contract with these types of issues. These errors are also often overlooked because they are embedded in otherwise innocuous contract language and surfaced only with both deep contextual awareness and exacting attention to detail.

Each category carries distinct legal implications. To illustrate, consider the `Uncapitalized Defined Terms'. Where a contracting party defines "Representative" narrowly, encompassing only company officers and legal counsel, but subsequently uses "representative" in lowercase within a confidentiality provision, a counterparty may reasonably interpret the inconsistency as intentional and apply the broader, ordinary meaning of the term. The result is that the countererparty may share sensitive information with a substantially wider group than the drafting party intended. By contrast, some errors may be characterized as more mundane in nature, with comparatively limited legal consequence in isolation. Nonetheless, even errors that appear minor can introduce ambiguity about agreed terms, generate friction between contracting parties, delay deal execution, and erode client confidence -- outcomes that carry real commercial cost regardless of their legal characterization. See Appendix \ref{app:implication} for implication of each category.

\paragraph{Construction Pipeline}

The process consists of three stages as shown in Figure~\ref{fig:pipeline}: (i) selecting and reviewing source contracts for structural flaws; (ii) annotating existing issues; and (iii) inserting additional issues to create the final gold answers. Source contracts are drawn from the open-source CUAD dataset \cite{hendrycks1cuad}, which is in turn drawn from EDGAR\footnote{https://www.sec.gov/search-filings}, an open repository of documents from publicly-owned US companies. \ourbenchmark~does \textbf{not} use CUAD's labels: each contract is independently reviewed and annotated by legal subject matter experts (SMEs) for categories relevant to scrubbing contracts. We describe the pipeline in greater detail:

\begin{enumerate}
    \item \textbf{Contract sourcing and review.} Contracts are selected from the source pool to span a range of subject matter and drafting styles. Each candidate is reviewed by an SME for fundamental drafting flaws that would render it unusable, and for suitable length - short enough to validate within reasonable effort, but long enough to host the full set of targeted issues without becoming structurally invalid, which in practice means approximately 10-15 pages. Where necessary, contracts were also revised to improve overall coherence by, for example, removing empty exhibits or improving consistency. This process yields an initial corpus $\mathcal{\hat{C}} = \{\hat{c}_1, \dots, \hat{c}_n\}$.

    \item \textbf{Annotation of existing issues.} For each $\hat{c}_i$, the SME records every defined term and any pre-existing instances of error categories $k \in \mathcal{K}$ already present in the contract as tuples $(\kappa, \mathbf{f})$. The quality of contracts in this dataset varies widely, and some contracts will have several existing errors to be identified.
    
    \item \textbf{Insertion of new issues.} The SME then purposefully introduces additional drafting errors across the error categories (Table \ref{tab:scrub_categories}), reflecting realistic mistakes seen in transactional practices. These annotations aim for broad and approximately balanced representation across error types $\mathcal{K}$, while preserving the coherence and legal plausibility of the edited contract. Each inserted issue is recorded as a tuple $(\kappa, \mathbf{f})$ under the same schema as pre-existing issues. Every occurrence is logged separately: repeated terms or errors in the same section are not de-duplicated. The final result is a new contract, $c_i \in \mathcal{C}$ for each original contract $\hat{c}_i$, and gold answers, $R_i$, including all defined terms, any identified pre-existing issues and all SME-inserted issues.
\end{enumerate}

After the data creation pipeline, two of the lawyers performed targeted reviews of the contracts and gold answers for quality, and suggested changes to the contracts, gold answers, and task prompts as necessary to increase alignment.

\subsection{Metrics}

We evaluate performance via multiset comparison between gold and predicted tuples. For each category $k \in \mathcal{K}$, tuples from $R_i$ and $\hat{R}_i$ are normalized and matched across all contracts; true positives ($TP_k$) are matched pairs, false positives ($FP_k$) are unmatched tuples in $\hat{R}_i$, and false negatives ($FN_k$) are unmatched tuples in $R_i$, with counts pooled across all $c_i \in \mathcal{C}$. 

We focus on recall, $R_k=\frac{TP_k}{TP_k+FN_k}$, as the primary metric, though we also report precision and F1 scores. For contract scrubbing, the potential costs of false negatives are significantly higher than the cost of false positives, since it is much easier to check whether a flagged issue is real than to identify issues that were not previously known. Practically, since the CUAD contracts can also contain pre-existing errors, using recall allows us to focus primarily on the known issues that experts have identified and inserted without being impacted by any potential remaining unknown issues.

Overall performance is reported as the macro-average across all 
$|\mathcal{K}| = 9$ categories:
\[
\small
    \mathrm{Macro\text{-}R} = \frac{1}{|\mathcal{K}|}\sum_{k \in \mathcal{K}} R_k.
\]
We additionally report a \textit{word-only} variant in which location fields are removed from each tuple prior to comparison. This isolates errors of \emph{identification} -- whether the model found the correct term, reference, or party pair -- from errors of \emph{localization}, i.e., whether it cited the correct section.

\subsection{Models}
We evaluate a range of proprietary and open-source models: GPT-5.5, GPT-5.2, o4-mini; Claude Opus 4.7, Claude Sonnet 4.6, Claude Haiku 4.5; Qwen 3.5-397B; and Gemini 3.1 pro, Gemini 2.5 pro, Gemma-4-26B. This selection spans multiple model families and capability tiers -- from frontier flagship models to smaller, efficient variants -- enabling us to assess how model scale and family affect performance on the
legal document scrubbing task. We provide a full list of the models tested and their hyperparameters in Appendix~\ref{app:models}.

\subsection{Implementation Details}
Each model is prompted to scrub a given contract, with instructions describing each of the nine categories and the output and reference formats prompt (See the full prompts in Appendix~\ref{app:prompt}). We prompt the models to identify issues from each category in separate instances to help reduce the competing task demands. Each response is expected to return a single JSON object with one key per
category $k \in \mathcal{K}$. Model outputs are parsed into category-specific tuples -- for example, $(\textit{term},
\textit{location})$ for defined terms and $(\textit{wrong reference},
\textit{correct reference}, \textit{location})$ for incorrect references. These tuples are compared deterministically against gold annotation $R_i$ as multisets, so repeated occurrences are scored independently as the same term or error type may occur multiple times in different locations, and each occurrence imposes a separate burden on a reviewer.

\paragraph{Normalisation before scoring}
To avoid penalizing superficial formatting differences, all tuple fields are normalized on both the gold and prediction sides before comparison. We apply normalization as follow: (1) \textit{Lower-casing of Terms: } Term and word fields are lowercased and stripped of whitespace (`Licensor' and `licensor' match). (2) \textit{Location field Canonicalization:} Location fields are canonicalized by collapsing parenthetical section levels while preserving major/minor dots: e.g., \texttt{1(a)(i)} $\rightarrow$ \texttt{1ai}, \texttt{1.1(h)(vii)} $\rightarrow$ \texttt{1.1hvii}, so \texttt{1.1(d)} $\neq$ \texttt{11(d)}. Special labels such as \texttt{P}/\texttt{Preamble}, \texttt{R}/\texttt{Recitals}, \texttt{Exhibit X}, \texttt{Schedule X}, and \texttt{Signature Block} are also canonicalized. (3) \textit{Symmetric Tuple Matching:} Paired-location categories, such as Conflicting Definitions and Inconsistent Terms, are compared order-independently: a gold tuple containing locations \texttt{(1e, 7a)} is treated as equivalent to a predicted tuple containing \texttt{(7a, 1e)}.

\section{Results}

\begin{wraptable}{l}{0.5\linewidth}
\centering
\small
\caption{Main Results. Precision (P), Recall (R), F1 scores, and cost for all evaluated models.
The best score in each column is \textbf{in bold}; the second best is \underline{underlined}. Cost column includes the mean price (\$) and time (sec.) per contract.}
\label{tab:main_results}
\renewcommand{\arraystretch}{1.2}
\begin{tabularx}{\linewidth}{p{0.3\linewidth}ccccc}
\toprule
\textbf{Model} & \textbf{R} & \textbf{P} & \textbf{F1} & \textbf{Cost (\$ | s)} \\
\midrule
GPT-5.5  & \textbf{0.750}& 0.580 & 0.632 & 1.38 | 533 \\
Gemini 3.1 Pro   & \underline{0.744}    & 0.616 & \textbf{0.655} & 0.19 | 76 \\
Claude Sonnet 4.6   & 0.686 & \underline{0.620} & \underline{0.637} & 1.52 | 258 \\
Gemini 2.5 Pro  & 0.632 & 0.527 & 0.557 & 0.13 | 48 \\
Claude Opus 4.7   & 0.616 & \textbf{0.644} & 0.621 & 0.66 | 23\\
GPT-5.2  & 0.589 & 0.540 & 0.553 & 0.32 | 192\\
Qwen 3.5 (397B) & 0.438 & 0.268 & 0.316 & 0.04 | 194 \\
Claude Haiku 4.5 & 0.445 & 0.592 & 0.492 & 0.68 | 108\\
o4-mini & 0.409 & 0.548 & 0.453 & 0.22 | 301 \\
\\

\midrule
\midrule
\multicolumn{5}{l}{\textit{\textbf{Without} Reasoning}} \\
\midrule
GPT-5.5  & 0.643 & 0.463 & 0.525 & 0.42 | 22\\
Claude Opus 4.7 & 0.526 & 0.478 & 0.490 & 0.42 | 5 \\
Gemma 4 (26B) & 0.365 & 0.315 & 0.326& 0.01 | 26 \\

\bottomrule
\end{tabularx}
\smallskip
\end{wraptable}

\paragraph{Overall} The benchmark is reasonably challenging, with GPT-5.5 having a recall score of 0.750, and the precision and F1 scores across models nearly all less than 0.650 as shown in Table~\ref{tab:main_results}. Most models have higher recall than precision, but there are tradeoffs between the two scores, with the model having the highest recall, GPT-5.5, actually having being fourth in precision. Focusing on recall, performance drops off quickly from the top few models to the weaker ones. The most expensive model, GPT-5.5, costs only \$1.38 per contract, significantly less than a lawyer or paralegal, though at scale the cost differences to other models may be more relevant. Runtime is also a consideration, with the top performance of GPT-5.5 also requiring the most latency, taking nearly 9 minutes to finish while Gemini 3.1 Pro returns within 90 seconds.

As expected, smaller and older models generally lag behind their larger counterparts. However, Qwen3.5-397b (.438 recall) performs poorly despite the large parameter count, comparable to the much smaller o4-mini (.409) and Haiku 4.5 (.445). This suggests that raw model scale does not straightforwardly translate to legal document analysis ability.

\begin{table*}[]
\centering
\caption{Recall score comparison across models and scrub categories.
\small\textit{Color heatmap applied to all metric rows. Recall Overall shown without heatmap for reference. All models use reasoning variants.}}
\resizebox{0.9\textwidth}{!}{%
\begin{tabular}{
  >{}l
  c  c  c  c  c  c  c  c  c 
}
\toprule
\multirow{3}{*}{\textbf{Metric}} &
  \multicolumn{9}{c}{\textbf{Model}} \\
\cmidrule(lr){2-10}
 & \multirow{2}{*}{\textbf{GPT-5.5}} & \textbf{Gemini} & \textbf{Claude} & \textbf{Gemini} & \textbf{Claude} & \multirow{2}{*}{\textbf{GPT-5.2}} & \textbf{Claude} & \textbf{Qwen3.5} & \multirow{2}{*}{\textbf{o4-mini}} \\[-4pt]
 & & \textbf{3.1 Pro} & \textbf{Sonnet 4.6} & \textbf{2.5 Pro} & \textbf{Opus 4.7} & & \textbf{Haiku 4.5} & \textbf{(397B)} & \\
\midrule
\rowcolor{gray!10}
\textbf{Overall Recall}
  & \textbf{0.750}
  & 0.744
  & 0.686
  & 0.632
  & 0.616
  & 0.589
  & 0.445
  & 0.438
  & 0.409
  \\\midrule
\addlinespace[2pt]
Defined Terms
  & \heatcell{0.905}
  & \heatcell{0.901}
  & \heatcell{0.862}
  & \heatcell{0.862}
  & \heatcell{0.835}
  & \heatcell{0.876}
  & \heatcell{0.753}
  & \heatcell{0.793}
  & \heatcell{0.730} 
  \\
Undef.\ Capitalized Terms
  & \heatcell{0.514}
  & \heatcell{0.440}
  & \heatcell{0.367}
  & \heatcell{0.319}
  & \heatcell{0.447}
  & \heatcell{0.543}
  & \heatcell{0.141}
  & \heatcell{0.216}
  & \heatcell{0.171} 
  \\
Uncapitalized Defined Terms
  & \heatcell{0.868}
  & \heatcell{0.868}
  & \heatcell{0.703}
  & \heatcell{0.662}
  & \heatcell{0.596}
  & \heatcell{0.628}
  & \heatcell{0.306}
  & \heatcell{0.331}
  & \heatcell{0.227} 
  \\
Incorr.\ Capitalized in Context
  & \heatcell{0.744}
  & \heatcell{0.690}
  & \heatcell{0.527}
  & \heatcell{0.357}
  & \heatcell{0.481}
  & \heatcell{0.419}
  & \heatcell{0.171}
  & \heatcell{0.349}
  & \heatcell{0.101} 
  \\
Unused Defined Terms
  & \heatcell{0.936}
  & \heatcell{0.931}
  & \heatcell{0.896}
  & \heatcell{0.782}
  & \heatcell{0.871}
  & \heatcell{0.822}
  & \heatcell{0.599}
  & \heatcell{0.485}
  & \heatcell{0.703} 
  \\
Terms Defined Multiple Times
  & \heatcell{0.753}
  & \heatcell{0.835}
  & \heatcell{0.825}
  & \heatcell{0.722}
  & \heatcell{0.763}
  & \heatcell{0.670}
  & \heatcell{0.526}
  & \heatcell{0.577}
  & \heatcell{0.526} 
  \\
Incorr.\ Sec./Art./Para.\ Refs
  & \heatcell{0.760}
  & \heatcell{0.800}
  & \heatcell{0.747}
  & \heatcell{0.733}
  & \heatcell{0.773}
  & \heatcell{0.667}
  & \heatcell{0.687}
  & \heatcell{0.533}
  & \heatcell{0.533}
  \\
Incorrect Party References
  & \heatcell{0.562}
  & \heatcell{0.569}
  & \heatcell{0.531}
  & \heatcell{0.569}
  & \heatcell{0.200}
  & \heatcell{0.008}
  & \heatcell{0.315}
  & \heatcell{0.200}
  & \heatcell{0.308}
  \\
Inconsistent Language
  & \heatcell{0.713}
  & \heatcell{0.667}
  & \heatcell{0.713}
  & \heatcell{0.678}
  & \heatcell{0.575}
  & \heatcell{0.667}
  & \heatcell{0.506}
  & \heatcell{0.460}
  & \heatcell{0.379} 
  \\
\bottomrule
\end{tabular}
}
\label{tab:category_table}
\end{table*}

\paragraph{Performance by Category}
As shown in Table \ref{tab:category_table}, model performance varies widely across the different scrub categories. The gap in recall scores between the easiest (Defined Terms) and hardest category (Undefined Capitalized Terms) is 0.484. Tasks which involve understanding legal relationships between different parts of the contracts tend to be more difficult than those which only require spotting repetitions or omissions. Easier categories -- Defined Terms (mean recall .835), Terms Defined Multiple Times (.689), Unused Defined Terms (.781) -- have explicit lexical signals and require less legal reasoning. In contrast, Incorrect Party References (.362), Incorrect Capitalization in Context (.427), and Undefined Capitalized Terms (.351) require inferring intent from context rather than matching surface form. Catching capitalized undefined terms requires the ability to flag inconsistencies against an internally maintained registry of definitions in combination with understanding which capitalizations are legally significant. Even the strongest performing model, GPT-5.5, achieves only 0.514 on Undefined Capitalized Terms and 0.562 on Party References.

\begin{figure*}
\begin{subfigure}[t]{0.4\linewidth}
    \centering
    \includegraphics[width=\linewidth]{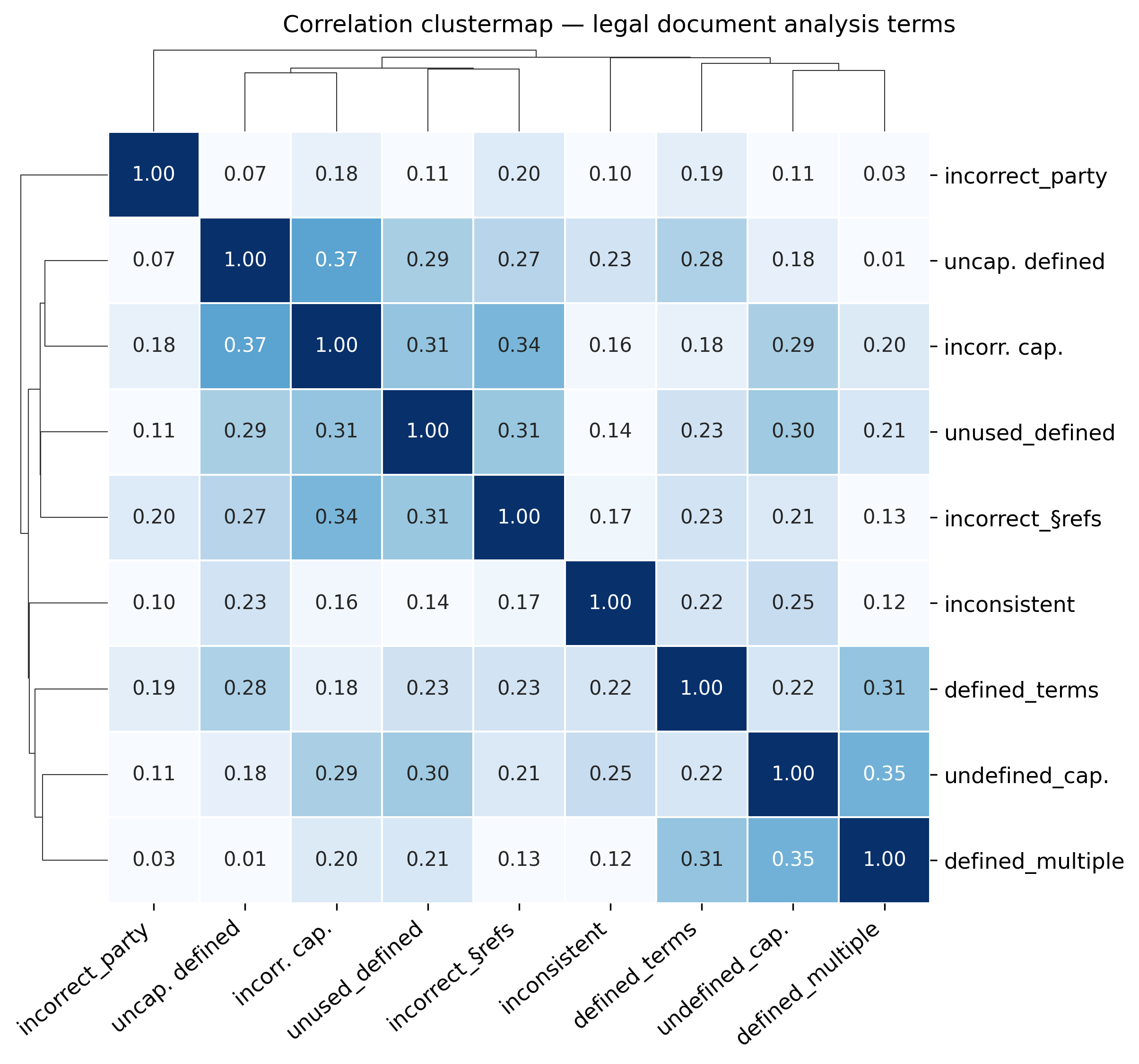}
    \caption{Correlation of Subtask Performance. Task categories show clusters, between those related to definitions, with less correlation between the other categories (Pearson R).}
    \label{fig:clustering}
\end{subfigure}\hfill
\begin{subfigure}[t]{0.55\linewidth}
    \centering
    \includegraphics[width=\linewidth]{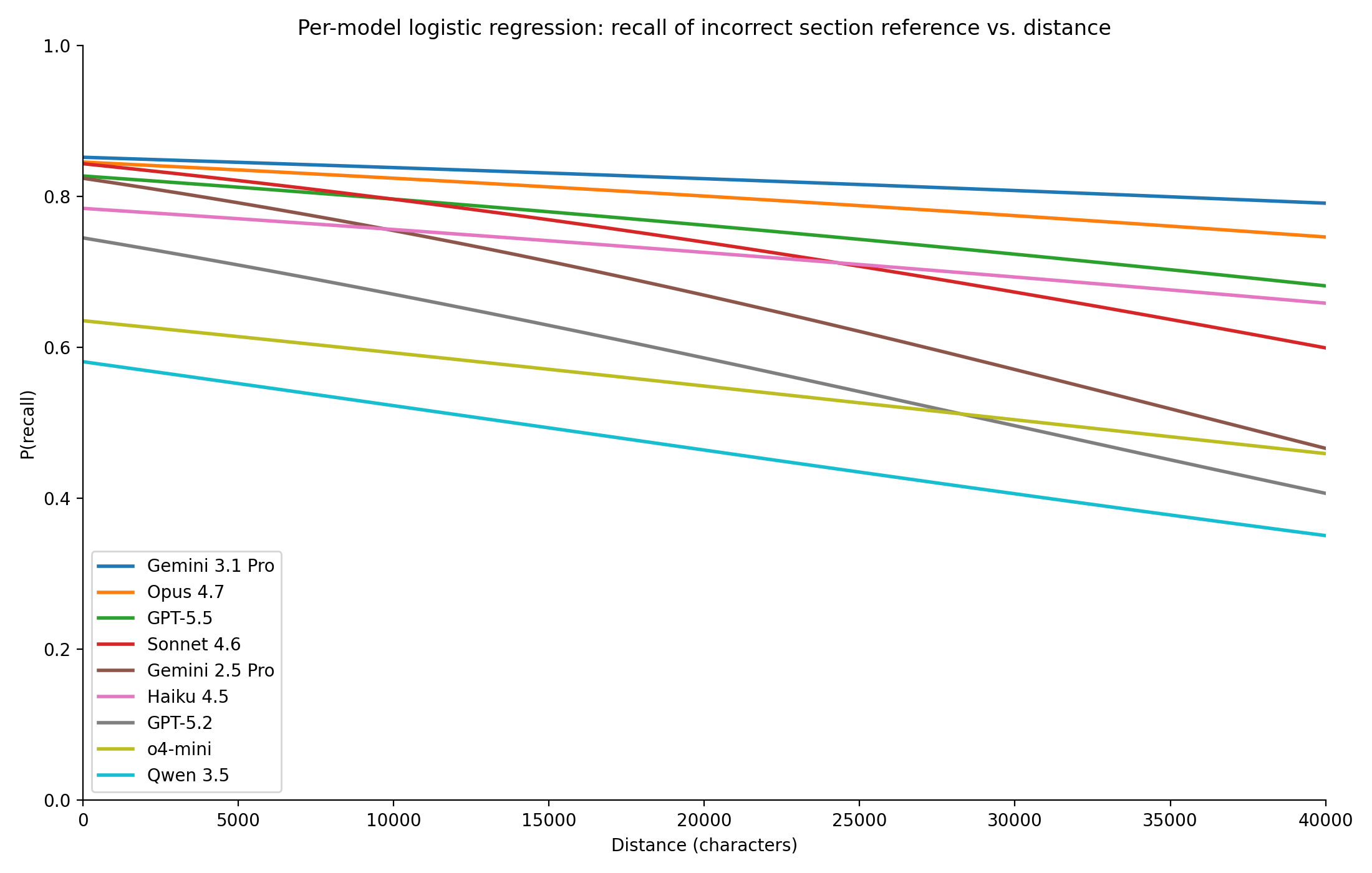}
    \caption{Long Distance References. Recall vs. reference distance, by model.}
    \label{fig:distance}
\end{subfigure}
\end{figure*}

Correlations (Pearson's R) between the category scores, shown in Figure~\ref{fig:clustering}, are also relatively low, but positive. There is a weak cluster of categories around incorrect usage of definitions, with uncapitalized defined terms, unused defined terms, incorrectly capitalized terms, and incorrect references forming a group, but overall the different subtasks appear to measure mostly independent capabilities.

\paragraph{Reasoning} As some of the tasks involved in contract scrubbing may appear not to require heavy reasoning, we tested whether turning off reasoning impacts model performance for the top models. The scores without reasoning are shown at the end of Table~\ref{tab:main_results}. For the two models we tested in both modes, including reasoning increased performance moderately at the cost of slower and more expensive inference. We include an analysis by issue category in Appendix~\ref{app:reasoning}.

\paragraph{Long-Distance Referencing} In reviewing the errors made by the models, we observed a qualitative tendency to struggle more with identifying incorrect section references that were further away (in terms of characters) from the section that they were meant to reference. As a targeted quantitative assessment, we manually annotated the distance in characters between the references and the referred sections for half of the ``Incorrect Section References'' gold standard items. As shown in Figure~\ref{fig:distance}, the general tendency is that predicted recall decreases as the distance between two target entities grows, with the effect becoming most pronounced past ten thousand characters (roughly 5-6 pages).
 

\section{Discussion}

\paragraph{Contract scrubbing in deployment} 
Contract scrubbing is a practical task requiring reasoning over long documents with close attention to detail. Although the mechanics of the task (e.g. spotting capitalization errors and checking for duplicative definitions) seem simple, legal knowledge and judgement are still an important aspect. Such foundational tasks, though routine, are also load-bearing: errors at this level propagate upward, undermining careful negotiations, expected outcomes, and client relationships. 
Full automation of contract scrubbing requires a high bar for model performance, likely to be higher than the 75\% recall currently attained. However, the speed and low cost of all of the evaluated models points to potential for integrating LLMs into existing workflows alongside experts, as well as adding more frequent scrubbing into contract lifecycles to reduce the issues that remain to be found at the end.

\textbf{Sources of task difficulty for LLMs}
The difficulty of contract scrubbing stems from the simultaneous demands of several core tasks: (1) \textit{long-context reasoning}, as errors must be detected across lengthy documents; (2) \textit{structured consistency checking}, which requires the model to maintain and query an implicit index of defined terms throughout the document; (3) \textit{sparse error detection}, where errors are rare, demanding high precision to avoid hallucination while retaining sufficient sensitivity to catch true positives; (4) \textit{heterogeneous error types}, meaning a single context includes various types of errors; and (5) \textit{document-internal context dependence}, where correctness is determined not by external legal standards but by conventions established within the document itself.

A number of existing benchmarks focus on measuring one or more of these abilities in more generic and isolated contexts, and no benchmark combines all of them, and the scores on our benchmark are lower than might be expected based on other existing benchmarks. For example, our inconsistent term detection category resembles the inconsistency detection evaluated in FaithEval~\cite{ming2025faitheval}, where models determine whether a passage contains internal contradictions. On FaithEval, GPT-4 and Claude Sonnet 3.5 achieved 89.4\% and 92.2\%, respectively, while more recent models from the same family achieve only 75.0\% and 68.6\% recall on inconsistency detection in \ourbenchmark. Similarly, our term-finding tasks are similar in concept to named entity recognition or needle in a haystack tasks, where models search for specific target words or phrases with long documents. \citeauthor{chen-etal-2025-longleader} evaluate retrieval from long contexts, where Claude 3 achieves 98.28\% at 128K tokens; Gemini 1.5 likewise reports near-perfect recall on needle-in-a-haystack probes \cite{google2024needle}. We do find in our results that performance across the tasks related to defined terms is correlated, but all of the tasks prove to be substantially harder in our setting than standard needle in a haystack. In addition to requiring many capabilities at once, we expect that the domain specific terminology and structure of legal contracts adds in a layer of difficulty further lowering scores.

\textbf{Broader implications}

Our findings in \ourbenchmark~highlight two important themes. First, they demonstrate a disconnect between perceived difficulty of a task within a professional domain and its difficulty for LLMs. This echoes observations that models can struggle on tasks that appear relatively simple \cite{nezhurina2024alice, dellacquaNavigatingJaggedTechnological2023}, even when they are capable of passing the SAT or the bar exam. Building on this, the gap between model performance on \ourbenchmark~and on existing benchmarks targeting individual capabilities suggests that evaluations of isolated abilities may not fully capture how models perform when those abilities are exercised jointly. Targeted benchmarks may focus on skills that seem cognitively significant to humans, while potentially overlooking factors that humans find trivial but which actually pose challenges for an LLM. Our analysis of the relationship between the distance between inconsistencies and model performance highlights one such example, where many models struggled with checking section references across longer spans. Contract scrubbing also requires following conventions internal to the context of a document rather than external world or domain-specific knowledge, which we believe is relatively under-represented in current benchmark efforts. 

\paragraph{Limitations}
\ourbenchmark~has a few limitations worth considering. First, while the benchmark has a large number of issues for the models to identify, they are drawn from only 44 contracts, a more modest scale. This size is sufficient to surface meaningful performance differences across models, but a larger corpus would reduce the effects of idiosyncrasies in any particular contract and better represent the universe of corporate contracts. Second, the benchmark's dataset contains only English-language contracts with an approximate length of 10--15 pages. This design choice reflects the practical setting of a closing-stage scrub in many real use cases, but limits generalisability to other legal traditions, languages, and deal types involving different document lengths such as shorter term sheets or multi-hundred-page complex financings. Third, our evaluation protocol requires models to produce structured JSON output, which may depress raw performance scores relative to a free-form setting and introduces sensitivity to instruction-following ability as a confounder. That said, structured output is arguably a realistic requirement for any production scrubbing tool, where downstream parsing and issue tracking depend on machine-readable responses; the restriction, therefore, reflects a genuine constraint of the deployment context rather than an arbitrary evaluation choice.

\section{Conclusion}
We introduce \ourbenchmark, a novel evaluation for assessing the ability of LLMs to scrub legal contracts for errors. We found that performance varied widely among models, with many scoring well on tasks that are primarily lexical in nature and scoring poorly when the tasks require more complex legal reasoning on top of contextual document analysis skills. These findings contrast with results on related general-domain benchmarks, where frontier models are generally very capable. Our results reinforce the case for narrowly targeted, ecologically valid, benchmarks in professional domains. As AI systems are increasingly deployed in professional settings, benchmarks that directly measure economically relevant, task-specific performance are important for both scientific progress and responsible deployment. We offer \ourbenchmark~as one step in that direction.



\clearpage
\printbibliography

\newpage
\appendix
\onecolumn

\noindent\textbf{\large Appendix}
\vspace{0.5em}

\noindent
\begin{tabularx}{\linewidth}{@{}lXr@{}}
\toprule
  \textbf{PART 1} & \textbf{Additional Context and Experiments}  \\
  A & Potential Implications of Each Error Category & \pageref{app:implication} \\
  B & Reasoning Ablation & \pageref{app:reasoning} \\
  C & Term-only Scoring Ablation & \pageref{app:scoring1} \\
  D & Per-category Precision \& Recall & \pageref{app:prec_f1} \\
  \midrule
  \textbf{PART 2} & \textbf{Reproducibility Details} \\
  E & Data Access and Usage Notes & \pageref{app:data} \\
  F & Inference Details & \pageref{app:models} \\
  G & Prompt Templates & \pageref{app:prompt} \\
  
\bottomrule
\end{tabularx}
\vspace{1em}

{\Large \textbf{PART 1: Additional Context and Experiments}}
\hrule

\section{Potential implications of each error category}
\vspace{-1em}
\begin{table}[h]
\caption{Scrub error category counts in \ourbenchmark~and per-contract averages}
\centering
\small
\begin{tabular}{lrr}
\toprule
\textbf{Category} & \textbf{Total} & \textbf{Avg/Contract} \\
\midrule
Defined Terms & 1,505 & 34.2 \\
Undefined Capitalized Terms & 689 & 15.7 \\
Uncapitalized Defined Terms & 317 & 7.2 \\
Unused Defined Terms & 202 & 4.6 \\
Incorrect Section, Article, or Paragraph References & 150 & 3.4 \\
Incorrect Party References & 130 & 3.0 \\
Incorrectly Capitalized Terms In Context & 129 & 2.9 \\
Terms Defined Multiple Times & 97 & 2.2 \\
Inconsistent Language & 87 & 2.0 \\
\bottomrule
\end{tabular}
\label{tab:category_counts}
\end{table}

\label{app:implication}
Each category has different potential implications for the parties agreeing to the contract. Not all consequences are strictly "legal" in nature such that a law is broken or a contract is breached. Courts may, in certain circumstances, correct obvious clerical or scrivener's errors where the parties' mutual intent is evident from the surrounding context. Nevertheless, such errors remain undesirable, and those that introduce substantive ambiguity are of greater legal concern. From a practical standpoint, however, the threshold for harm is lower. Any error that delays deal execution or introduces friction between parties, including ostensibly minor clerical mistakes that create confusion about the operative terms, represents a failure with real professional consequences, up to and including reputational damage and client attrition.

Below, we describe what each category represents and provide examples of possible impacts that could arise from missing this type of error:

\begin{table}[h]

\begin{tabular}{p{0.95\columnwidth}}
    \textbf{Undefined Capitalized Terms}:
Terms that are capitalized or otherwise treated as a defined term in an agreement but not formally defined. For example, an agreement provides: ``Party A shall comply with all applicable requirements in the Approved Specification,'' but Approved Specification is not defined. This matters because an undefined capitalized term indicates that a contracting party specifically intended a special contractual meaning. As a result, an ambiguity is created and a party may later dispute what specification was approved and whether it was applicable, whether a breach occurred, and whether outside evidence is admissible to contradict, vary, or add to the terms of the agreement. \\

\end{tabular}
\end{table}
\begin{tabular}{p{0.95\columnwidth}}

\textbf{Uncapitalized Defined Terms} Formally defined terms appearing in lowercase when they should be capitalized. For example, defining ``Representative'' to mean only company officers and legal counsel, rather than the everyday meaning of anyone acting on another's behalf. If the confidentiality section of an agreement then says that a party can disclose confidential information to ``representatives'' in lowercase, a counterparty may treat the inconsistency as intentional and apply the broader, everyday definition. The result is that the other party could be permitted to share sensitive information with a much wider group of people than was intended.\\
\textbf{Incorrectly Capitalized Terms in Context}
Terms that have a definition but are capitalized in a context where they are not being used as a defined term. For example, an agreement provides: ``\textquoteleft Services\textquoteright\ shall mean the software implementation services provided by Vendor to Customer as described in Exhibit A.'' The agreement later provides that: ``Vendor shall not provide similar Services to any competitor of Customer.'' This matters because the later capitalized term can change the scope of a covenant, restriction, exclusion, or permission. Here, the later erroneous capitalization improperly restricts what should be ``services'' to the implementation services in Exhibit A whereas a drafter could have intended a broader scope of similar services generally provided by the Vendor.\\

\textbf{Unused Defined Terms}
Defined terms that appear only once in the context where they are defined and are not used elsewhere in the agreement. For example, an agreement provides: ``\textquoteleft Change Order\textquoteright\ means additional or different specifications from the project terms set out in Scope of Construction.'' However, the term ``Change Order'' never appears again in the agreement. Because courts aim to avoid any interpretation of contractual language that renders it surplusage, disputes over intent and scope of additional or different specifications can arise if a party attributes meaning to the stranded definition and can invite an opportunity to seek the admission of extrinsic evidence to provide competing interpretations of the agreement.\\

\textbf{Terms Defined Multiple Times}
Terms that are defined more than one time in an agreement with conflicting or inconsistent definitions. A term that is merely repeated with the same meaning is NOT an error --- only flag a term when its multiple definitions genuinely conflict. For example, if the Effective Date is defined in two places with conflicting dates in an agreement, the parties may disagree about which one controls. Consider a company hired to manage a property that is liable for any accidents occurring while the agreement is in effect. If the agreement defines the Effective Date as both March 1 and March 30, and an accident occurs on March 15, it becomes genuinely unclear who is responsible for that incident, the property owner or the management company. What should be a straightforward question can turn into a costly dispute.\\

\textbf{Incorrect Section, Article, or Paragraph References} Internal cross-references that assign an incorrect section, article, or paragraph of the agreement. Agreements are often reorganized during negotiation, with sections added, removed, or reordered. If an internal reference isn't updated to reflect those changes, it may end up pointing to the wrong section entirely. For example, suppose an agreement states that a party will face enhanced damages for breaches of ``Section 3,'' which covers confidentiality. During negotiation, the sections are reshuffled and Section 3 now covers product liability instead, but the reference is never updated. That party is now potentially exposed to enhanced damages for product liability incidents, a potentially broader and more expensive risk.\\

\textbf{Incorrect Party References}
Instances where a party is referred to by the wrong party name or role, such as ``Licensor'' instead of ``Licensee'' or ``Receiving Party'' instead of ``Disclosing Party''. Referencing the wrong party name in a contract can shift responsibilities in ways that a party may not want. If a clause requires a specific action, such as paying for shipping, but names the wrong party, the obligation could fall on that party regardless of what was originally discussed during contract negotiations. This type of error can end up costing a party time and money and potentially lead to a dispute over who is actually responsible.\\

\textbf{Inconsistent Language}
Language in the agreement that directly contradicts itself or other language elsewhere in the agreement. When two provisions in a contract directly contradict each other, it creates uncertainty about which one actually applies. For example, if one section states that payment is due 30 days after receiving an invoice, but another states 45 days, neither party can be fully confident about when payment is expected. For the seller, this kind of ambiguity can complicate cash flow planning; they may be counting on payment at 30 days, while the buyer believes they have until 45. What starts as a drafting oversight can quickly become a source of friction or dispute.\\
\end{tabular}

\begin{figure*}
    \centering
    \includegraphics[width=1\linewidth]{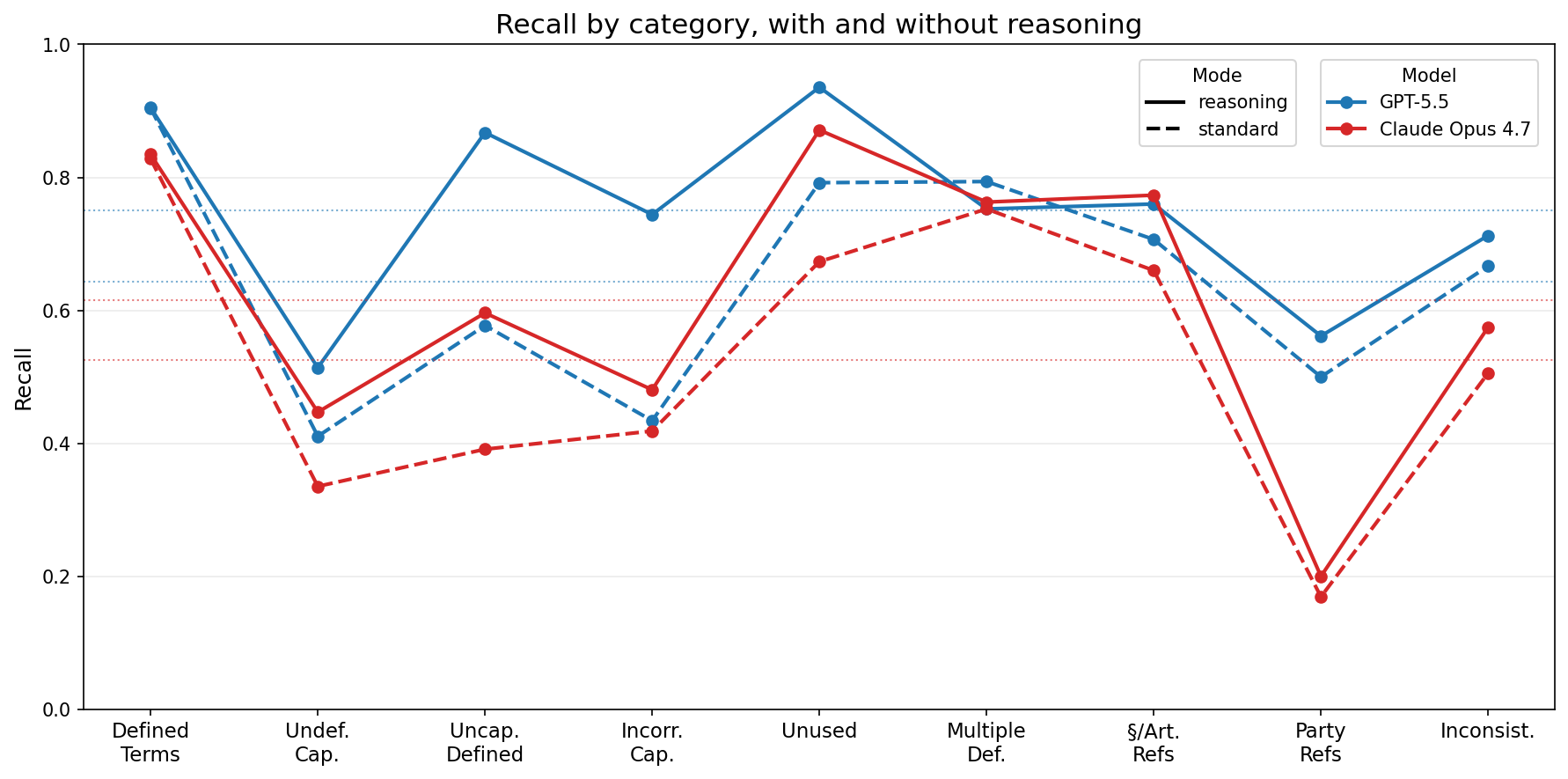}
    \caption{Recall by categories for two models with and without reasoning. Solid lines show per-section recall across nine defined-term and reference-checking tasks; dashed horizontal lines indicate each condition's mean.}
    \label{fig:reasoning_analysis}
\end{figure*}

\section{Reasoning Ablation}
\label{app:reasoning}
As some of the tasks involved in contract scrubbing may appear not to require heavy thinking, we tested whether turning off reasoning impacts model performance for the top models. The scores with and without reasoning are shown in Figure \ref{fig:reasoning_analysis}.

Across the two models tested, enabling reasoning markedly improved performance for Claude Opus 4.7 (+0.090 avg. recall) and GPT-5.5 (+0.107). The effect was particularly pronounced for Unused Defined Term category and Uncapitlized Defined Term categories. Although these categories rely on lexical signals rather than deep legal interpretation, verifying them requires holistic reasoning over the full document rather than local pattern-matching, which likely explains why reasoning-enabled variants benefit more.
By contrast, categories that depend on inferring intent, for example, Incorrect Party References, saw little to no improvement from reasoning, suggesting that current reasoning traces help with structured cross-referencing more than with the deeper semantic judgments these categories demand in this task. For this reason, we report the main results with reasoning enabled.

\section{Term-only Scoring Ablation}
\label{app:scoring1}

To distinguish between failure to identify issues in each contract and failure to provide their locations, we conducted a scoring ablation. Rather than multiset comparison across tuples from $R_i$ and $\hat{R}_i$, we limit the comparison to the categories, $\kappa$, and the elements of $\mathbf{f}$ which are terms, allowing mismatches in the locations. 

By construction, scores matching only on the terms are higher than the scores requiring both term and location matches (Table~\ref{tab:abl_term_only}). The differences are mostly on the order of five percentage points, though Qwen 3.5 sees much larger improvements and actually passes Claude 4.5 Haiku, indicating that it struggles more with location references within the document than the other models.

\begin{table*}[]
\centering
\caption{Recall score comparison across models and scrub categories.
\small\textit{Color heatmap applied to all metric rows. Recall Overall shown without heatmap for reference. All models use reasoning variants.}}
\resizebox{0.9\textwidth}{!}{%
\begin{tabular}{
  >{}l
  c  c  c  c  c  c  c  c  c 
}
\toprule
\multirow{3}{*}{\textbf{Metric}} &
  \multicolumn{9}{c}{\textbf{Model}} \\
\cmidrule(lr){2-10}
 & \multirow{2}{*}{\textbf{GPT-5.5}} & \textbf{Gemini} & \textbf{Claude} & \textbf{Gemini} & \textbf{Claude} & \multirow{2}{*}{\textbf{GPT-5.2}} & \textbf{Claude} & \textbf{Qwen3.5} & \multirow{2}{*}{\textbf{o4-mini}} \\[-4pt]
 & & \textbf{3.1 Pro} & \textbf{Sonnet 4.6 \ddag} & \textbf{2.5 Pro} & \textbf{Opus 4.7} & & \textbf{Haiku 4.5} & \textbf{(397B)} & \\
\midrule
\rowcolor{gray!10}
\textbf{Overall Recall}
  & \textbf{0.799}
  & 0.793
  & 0.746
  & 0.698
  & 0.661
  & 0.631
  & 0.478
  & 0.529
  & 0.466
  \\\midrule
\addlinespace[2pt]
Defined Terms
  & \heatcell{0.940}
  & \heatcell{0.941}
  & \heatcell{0.919}
  & \heatcell{0.944}
  & \heatcell{0.890}
  & \heatcell{0.933}
  & \heatcell{0.809}
  & \heatcell{0.877}
  & \heatcell{0.791} 
  \\
Undef.\ Capitalized Terms
  & \heatcell{0.552}
  & \heatcell{0.460}
  & \heatcell{0.392}
  & \heatcell{0.347}
  & \heatcell{0.505}
  & \heatcell{0.582}
  & \heatcell{0.165}
  & \heatcell{0.253}
  & \heatcell{0.205}
  \\
Uncapitalized Defined Terms
  & \heatcell{0.924}
  & \heatcell{0.915}
  & \heatcell{0.776}
  & \heatcell{0.776}
  & \heatcell{0.656}
  & \heatcell{0.707}
  & \heatcell{0.369}
  & \heatcell{0.533}
  & \heatcell{0.297}
  \\
Incorr.\ Capitalized in Context
  & \heatcell{0.798}
  & \heatcell{0.736}
  & \heatcell{0.620}
  & \heatcell{0.395}
  & \heatcell{0.519}
  & \heatcell{0.481}
  & \heatcell{0.217}
  & \heatcell{0.519}
  & \heatcell{0.140} 
  \\
Unused Defined Terms
  & \heatcell{0.960}
  & \heatcell{0.955}
  & \heatcell{0.941}
  & \heatcell{0.837}
  & \heatcell{0.906}
  & \heatcell{0.876}
  & \heatcell{0.629}
  & \heatcell{0.540}
  & \heatcell{0.772} 
  \\
Terms Defined Multiple Times
  & \heatcell{0.794}
  & \heatcell{0.845}
  & \heatcell{0.866}
  & \heatcell{0.794}
  & \heatcell{0.794}
  & \heatcell{0.753}
  & \heatcell{0.588}
  & \heatcell{0.680}
  & \heatcell{0.608} 
  \\
Incorr.\ Sec./Art./Para.\ Refs
  & \heatcell{0.793}
  & \heatcell{0.840}
  & \heatcell{0.820}
  & \heatcell{0.813}
  & \heatcell{0.807}
  & \heatcell{0.713}
  & \heatcell{0.727}
  & \heatcell{0.573}
  & \heatcell{0.580}
  \\
Incorrect Party References
  & \heatcell{0.631}
  & \heatcell{0.654}
  & \heatcell{0.631}
  & \heatcell{0.677}
  & \heatcell{0.208}
  & \heatcell{0.008}
  & \heatcell{0.323}
  & \heatcell{0.254}
  & \heatcell{0.338}
  \\
\bottomrule
\end{tabular}
}
\label{tab:abl_term_only}
\end{table*}

\newpage
\section{Per-category Precision \& F1}
\label{app:prec_f1}
We provide per-categoy precision and F1 performance of all models we tested for completeness.

\subsection{Per-category Precision}
\begin{table*}[h]
\centering
\caption{Precision score comparison across models and scrub categories.
\small\textit{Color heatmap applied to all metric rows. Precision Overall shown without heatmap for reference.}}
\resizebox{\textwidth}{!}{%
\begin{tabular}{
  >{}l
  c  c  c  c  c  c  c  c  c 
}
\toprule
\multirow{3}{*}{\textbf{Metric}} &
  \multicolumn{9}{c}{\textbf{Model}} \\
\cmidrule(lr){2-10}
 & \multirow{2}{*}{\textbf{GPT-5.5}} & \textbf{Gemini} & \textbf{Claude} & \textbf{Gemini} & \textbf{Claude} & \multirow{2}{*}{\textbf{GPT-5.2}} & \textbf{Claude} & \textbf{Qwen3.5} & \multirow{2}{*}{\textbf{o4-mini}} \\[-4pt]
 & & \textbf{3.1 Pro} & \textbf{Sonnet 4.6 \ddag} & \textbf{2.5 Pro} & \textbf{Opus 4.7} & & \textbf{Haiku 4.5} & \textbf{(397B)} & \\
\midrule
\rowcolor{gray!10}
\textbf{Overall Precision}
  & 0.580
  & 0.616
  & 0.620
  & 0.527
  & 0.644
  & 0.540
  & 0.592
  & 0.268
  & 0.548
  \\\midrule
\addlinespace[2pt]
Defined Terms
  & \heatcell{0.878}
  & \heatcell{0.909}
  & \heatcell{0.901}
  & \heatcell{0.860}
  & \heatcell{0.916}
  & \heatcell{0.899}
  & \heatcell{0.890}
  & \heatcell{0.575}
  & \heatcell{0.906} 
  \\
Undef.\ Capitalized Terms
  & \heatcell{0.632}
  & \heatcell{0.596}
  & \heatcell{0.634}
  & \heatcell{0.564}
  & \heatcell{0.615}
  & \heatcell{0.570}
  & \heatcell{0.618}
  & \heatcell{0.185}
  & \heatcell{0.608}
  \\
Uncapitalized Defined Terms
  & \heatcell{0.327}
  & \heatcell{0.299}
  & \heatcell{0.408}
  & \heatcell{0.325}
  & \heatcell{0.543}
  & \heatcell{0.375}
  & \heatcell{0.406}
  & \heatcell{0.073}
  & \heatcell{0.369}
  \\
Incorr.\ Capitalized in Context
  & \heatcell{0.490}
  & \heatcell{0.556}
  & \heatcell{0.523}
  & \heatcell{0.354}
  & \heatcell{0.705}
  & \heatcell{0.470}
  & \heatcell{0.355}
  & \heatcell{0.019}
  & \heatcell{0.232} 
  \\
Unused Defined Terms
  & \heatcell{0.747}
  & \heatcell{0.752}
  & \heatcell{0.670}
  & \heatcell{0.640}
  & \heatcell{0.804}
  & \heatcell{0.731}
  & \heatcell{0.665}
  & \heatcell{0.359}
  & \heatcell{0.582} 
  \\
Terms Defined Multiple Times
  & \heatcell{0.777}
  & \heatcell{0.871}
  & \heatcell{0.816}
  & \heatcell{0.673}
  & \heatcell{0.841}
  & \heatcell{0.765}
  & \heatcell{0.797}
  & \heatcell{0.463}
  & \heatcell{0.797} 
  \\
Incorr.\ Sec./Art./Para.\ Refs
  & \heatcell{0.745}
  & \heatcell{0.682}
  & \heatcell{0.757}
  & \heatcell{0.611}
  & \heatcell{0.829}
  & \heatcell{0.725}
  & \heatcell{0.786}
  & \heatcell{0.362}
  & \heatcell{0.734} 
  \\
Incorrect Party References
  & \heatcell{0.376}
  & \heatcell{0.548}
  & \heatcell{0.496}
  & \heatcell{0.443}
  & \heatcell{0.228}
  & \heatcell{0.007}
  & \heatcell{0.423}
  & \heatcell{0.170}
  & \heatcell{0.377} 
  \\
Inconsistent Language
  & \heatcell{0.246}
  & \heatcell{0.331}
  & \heatcell{0.378}
  & \heatcell{0.269}
  & \heatcell{0.312}
  & \heatcell{0.314}
  & \heatcell{0.393}
  & \heatcell{0.207}
  & \heatcell{0.327}
  \\
\bottomrule
\end{tabular}
}
\label{tab:precision}
\end{table*}

\subsection{Per-category F1}
\begin{table*}[h]
\centering
\caption{F1 score comparison across models and scrub categories.
\small\textit{Color heatmap applied to all metric rows. F1 Overall shown without heatmap for reference.}}
\resizebox{\textwidth}{!}{%
\begin{tabular}{
  >{}l
  c  c  c  c  c  c  c  c  c 
}
\toprule
\multirow{3}{*}{\textbf{Metric}} &
  \multicolumn{9}{c}{\textbf{Model}} \\
\cmidrule(lr){2-10}
 & \multirow{2}{*}{\textbf{GPT-5.5}} & \textbf{Gemini} & \textbf{Claude} & \textbf{Gemini} & \textbf{Claude} & \multirow{2}{*}{\textbf{GPT-5.2}} & \textbf{Claude} & \textbf{Qwen3.5} & \multirow{2}{*}{\textbf{o4-mini}} \\[-4pt]
 & & \textbf{3.1 Pro} & \textbf{Sonnet 4.6 \ddag} & \textbf{2.5 Pro} & \textbf{Opus 4.7} & & \textbf{Haiku 4.5} & \textbf{(397B)} & \\
\midrule
\rowcolor{gray!10}
\textbf{Overall F1}
  & 0.632
  & \textbf{0.655}
  & 0.637
  & 0.557
  & 0.621
  & 0.553
  & 0.492
  & 0.316
  & 0.453
  \\\midrule
\addlinespace[2pt]
Defined Terms
  & \heatcell{0.891}
  & \heatcell{0.905}
  & \heatcell{0.881}
  & \heatcell{0.861}
  & \heatcell{0.873}
  & \heatcell{0.888}
  & \heatcell{0.816}
  & \heatcell{0.667}
  & \heatcell{0.809} 
  \\
Undef.\ Capitalized Terms
  & \heatcell{0.567}
  & \heatcell{0.506}
  & \heatcell{0.465}
  & \heatcell{0.408}
  & \heatcell{0.518}
  & \heatcell{0.556}
  & \heatcell{0.229}
  & \heatcell{0.199}
  & \heatcell{0.267} 
  \\
Uncapitalized Defined Terms
  & \heatcell{0.475}
  & \heatcell{0.445}
  & \heatcell{0.517}
  & \heatcell{0.436}
  & \heatcell{0.568}
  & \heatcell{0.470}
  & \heatcell{0.349}
  & \heatcell{0.119}
  & \heatcell{0.281}
  \\
Incorr.\ Capitalized in Context
  & \heatcell{0.591}
  & \heatcell{0.616}
  & \heatcell{0.525}
  & \heatcell{0.355}
  & \heatcell{0.571}
  & \heatcell{0.443}
  & \heatcell{0.230}
  & \heatcell{0.036}
  & \heatcell{0.141} 
  \\
Unused Defined Terms
  & \heatcell{0.831}
  & \heatcell{0.832}
  & \heatcell{0.767}
  & \heatcell{0.704}
  & \heatcell{0.836}
  & \heatcell{0.774}
  & \heatcell{0.630}
  & \heatcell{0.413}
  & \heatcell{0.637} 
  \\
Terms Defined Multiple Times
  & \heatcell{0.764}
  & \heatcell{0.853}
  & \heatcell{0.821}
  & \heatcell{0.697}
  & \heatcell{0.800}
  & \heatcell{0.714}
  & \heatcell{0.634}
  & \heatcell{0.514}
  & \heatcell{0.634} 
  \\
Incorr.\ Sec./Art./Para.\ Refs
  & \heatcell{0.752}
  & \heatcell{0.736}
  & \heatcell{0.752}
  & \heatcell{0.667}
  & \heatcell{0.800}
  & \heatcell{0.694}
  & \heatcell{0.733}
  & \heatcell{0.431}
  & \heatcell{0.618} 
  \\
Incorrect Party References
  & \heatcell{0.451}
  & \heatcell{0.558}
  & \heatcell{0.513}
  & \heatcell{0.498}
  & \heatcell{0.213}
  & \heatcell{0.008}
  & \heatcell{0.361}
  & \heatcell{0.184}
  & \heatcell{0.339} 
  \\
Inconsistent Language
  & \heatcell{0.366}
  & \heatcell{0.443}
  & \heatcell{0.494}
  & \heatcell{0.386}
  & \heatcell{0.405}
  & \heatcell{0.426}
  & \heatcell{0.442}
  & \heatcell{0.286}
  & \heatcell{0.351} 
  \\
\bottomrule
\end{tabular}
}
\label{tab:f1}
\end{table*}

\newpage
{\Large \textbf{PART 2: Reproducibility Details}}
\hrule

\section{Data Access and Usage Notes}
\label{app:data}
The dataset will be made publicly available upon acceptance. The CUAD dataset is used under a CC-BY-4.0 license.

\section{Inference Details}
\label{app:models}

We evaluated 10 models on the contract scrub benchmark. Details regarding hyperparameters and compute resources are listed in Table~\ref{tab:inference}. Empty cells indicate that an option is not available for a particular model.

\begin{table}[h]
    \centering
    \caption{Model Inference Details}
    \label{tab:inference}
    \begin{tabular}{cccccc}\toprule
         Name & Provider  & Temperature & Reasoning Effort\\
         \hline
         Claude Opus 4.7 & Bedrock & -- & High \\
         Claude Sonnet 4.6 & Bedrock & 0.6 & High \\
         Claude Haiku 4.5 & Bedrock & 0.6 & High \\
         GPT 5.5 & OpenAI & -- & Medium \\
         GPT 5.2 & OpenAI & -- & Medium \\
         o4 mini & Azure & 0.6 & -- \\
         Qwen 3.5 397B A17B & Self-hosted & 0.6 & -- \\
         Gemini 3.1 Pro & Vertex AI & 0.6 & -- \\
         Gemini 2.5 Pro & Vertex AI & 0.6 & -- \\
         Gemma 4 26B A4B & Vertex AI & 0.6 & -- \\\bottomrule
         
    \end{tabular}
\end{table}

\section{Prompt Templates}
\label{app:prompt}
We provide the full prompt text used to test each model in the main results. The prompts for each of the nine categories were built from an overall prompt template, with a targeted section inserted for each category. Each category has four items, including `title', `definition', `instruction' and `schema'.

\begin{tcolorbox}[colback=orange!5!white, colframe=orange!75!black,
                  title=Overall Scoring Prompt Template,
                  boxsep=2pt, left=4pt, right=4pt, top=2pt, bottom=2pt,
                  breakable]

You are a legal editor reviewing an agreement for drafting errors and issues.\\

Review the agreement below and identify ONLY the following category:\\

\{\texttt{title}\} - \{\texttt{definition}\}

\{\texttt{instructions}\}
\\\\
If you find no items in this category, respond with an empty list.
\\\\
Follow these rules for the location format. This is critical:

- Always give the most specific sub-section reference, NEVER just the parent section number.
\\\\
- Concatenate sub-section letters/numerals together; preserve dots between major.minor section numbers.
\\\\
- If a defined term appears in section 1(a), the location is "1a", NOT "1".
\\\\
- If something is in section 11(b)(i), the location is "11bi", NOT "11" or "11b".
\\\\
- If something is in section 7(a)(i), the location is "7ai", NOT "7" or "7a".
\\\\
- If something is in Attachment X Section Y, the location is "attachment xy", NOT "attachment x.y".
\\\\
- For nested numbering: section 1.1(d) is "1.1d", section 3.7(a) is "3.7a", section 1.1(h)(vii) is "1.1hvii".
\\\\
- Use "P" for the Preamble (the introductory section that identifies the parties, date and general background).
\\\\
- Use "Recitals" for the WHEREAS clauses; if the item is in a specific recital clause, append its letter/number, e.g. "Recitals c" for recital (c).
\\\\
- Use "Exhibit A", "Exhibit B" and "Schedule A", "Schedule B" etc. for exhibits and schedules.
\\\\
- Use "Signature Block" for the section where parties execute the agreement.
\\\\

Respond with JSON only - no explanation, no markdown fences. Use this exact schema:
\\\\
\{\texttt{schema}\}

\end{tcolorbox}

\begin{tcolorbox}[colback=orange!5!white, colframe=orange!85!white,
                  title= Defined Terms,
                  boxsep=2pt, left=4pt, right=4pt, top=2pt, bottom=2pt,
                  breakable]

\texttt{title:} Defined Terms\\

\texttt{definition:} Words or phrases that are formally defined in the agreement with a designated meaning within the context of the agreement, often set out as a word or words that are capitalized, underlined, bolded, italicized, or in quotations. \\

\texttt{instruction:} For each term, provide the section where it is defined. \\

\texttt{schema:} \{"defined\_terms": [\{"term": "...", "location": "..."\}]\}

\end{tcolorbox}

\begin{tcolorbox}[colback=orange!5!white, colframe=orange!85!white,
                  title= Undefined Capitalized Terms,
                  boxsep=2pt, left=4pt, right=4pt, top=2pt, bottom=2pt,
                  breakable]
\texttt{title:} Undefined Capitalized Terms\\

\texttt{definition:} Terms that are capitalized or otherwise treated as a defined term in an agreement but not formally defined. For example, an agreement provides: ``Party A shall comply with all applicable requirements in the Approved Specification,'' but Approved Specification is not defined. This matters because an undefined capitalized term indicates that a contracting party specifically intended a special contractual meaning. As a result, an ambiguity is created and a party may later dispute what specification was approved and whether it was applicable, whether a breach occurred, and whether outside evidence is admissible to contradict, vary, or add to the terms of the agreement.

However, the following shall NOT be flagged as undefined capitalized terms:
\begin{enumerate}
    \item Terms of art --- words or terms with a specific, precise, technical, and specialized meaning commonly understood and used within a particular field, profession, or discipline that is germane or applicable to the contract.
    \item Proper nouns used in common parlance --- names of specific people, places, brands, or things that have been adopted into everyday, informal speech and are widely understood by the general population, even outside of their original specialized or formal context.
    \item Morphological variants of defined terms --- instances where a defined term appears in a grammatically inflected form that is not itself defined.
    \item Scientific and technical nomenclature --- standardized words or terms with a specific, precise, and specialized meaning commonly understood and used in scientific and technical fields.
    \item References to agreement sections, exhibits, and schedules.
    \item The title of the agreement and the names of any associated documents as they appear in the agreement's header or title section.
    \item Titles, roles, and positions.
\end{enumerate}

\texttt{instruction:} List each occurrence separately with the section where each term is located. \\\\
\texttt{schema:} \{"undefined\_capitalized\_terms": [\{"term": "...", "location": "..."\}]\}
\end{tcolorbox}

\begin{tcolorbox}[colback=orange!5!white, colframe=orange!85!white,
                  title= Uncapitalized Defined Terms,
                  boxsep=2pt, left=4pt, right=4pt, top=2pt, bottom=2pt,
                  breakable]
\texttt{title:} Uncapitalized Defined Terms\\\\
\texttt{definition:} Formally defined terms appearing in lowercase when they should be capitalized. For example, defining ``Representative'' to mean only company officers and legal counsel, rather than the everyday meaning of anyone acting on another's behalf. If the confidentiality section of an agreement then says that a party can disclose confidential information to ``representatives'' in lowercase, a counterparty may treat the inconsistency as intentional and apply the broader, everyday definition. The result is that the other party could be permitted to share sensitive information with a much wider group of people than was intended. \\\\
\texttt{instruction:} List each occurrence separately with the section where each term is located. \\\\
\texttt{schema:} \{"uncapitalized\_defined\_terms": [\{"term": "...", "location": "..."\}]\}

\end{tcolorbox}

\begin{tcolorbox}[colback=orange!5!white, colframe=orange!85!white,
                  title= Incorrectly Capitalized Terms In Context,
                  boxsep=2pt, left=4pt, right=4pt, top=2pt, bottom=2pt,
                  breakable]

\texttt{title:} Incorrectly Capitalized Terms In Context\\\\
\texttt{definition:} Terms that have a definition but are capitalized in a context where they are not being used as a defined term. For example, an agreement provides: ``\textquoteleft Services\textquoteright\ shall mean the software implementation services provided by Vendor to Customer as described in Exhibit A.'' The agreement later provides that: ``Vendor shall not provide similar Services to any competitor of Customer.'' This matters because the later capitalized term can change the scope of a covenant, restriction, exclusion, or permission. Here, the later erroneous capitalization improperly restricts what should be ``services'' to the implementation services in Exhibit A whereas a drafter could have intended a broader scope of similar services generally provided by the Vendor. \\\\
\texttt{instruction:} List each occurrence separately with the section where each term is located. \\\\
\texttt{schema:} \{"incorrectly\_capitalized\_terms\_in\_context": [\{"term": "...", "location": "..."\}]\}

\end{tcolorbox}

\begin{tcolorbox}[colback=orange!5!white, colframe=orange!85!white,
                  title= Unused Defined Terms,
                  boxsep=2pt, left=4pt, right=4pt, top=2pt, bottom=2pt,
                  breakable]

\texttt{title:} Unused Defined Terms\\\\
\texttt{definition:} Defined terms that appear only once in the context where they are defined and are not used elsewhere in the agreement. For example, an agreement provides: ``\textquoteleft Change Order\textquoteright\ means additional or different specifications from the project terms set out in Scope of Construction.'' However, the term ``Change Order'' never appears again in the agreement. Because courts aim to avoid any interpretation of contractual language that renders it surplusage, disputes over intent and scope of additional or different specifications can arise if a party attributes meaning to the stranded definition and can invite an opportunity to seek the admission of extrinsic evidence to provide competing interpretations of the agreement. \\\\
\texttt{instruction:} List each occurrence separately with the section where each term is located. \\\\
\texttt{schema:} \texttt{\{"unused\_defined\_terms": [\{"term": "...", "location": "..."\}]\}}

\end{tcolorbox}

\begin{tcolorbox}[colback=orange!5!white, colframe=orange!85!white,
                  title= Terms Defined Multiple Times,
                  boxsep=2pt, left=4pt, right=4pt, top=2pt, bottom=2pt,
                  breakable]
\texttt{title:} Terms Defined Multiple Times\\\\
\texttt{definition:} Terms that are defined more than one time in an agreement with conflicting or inconsistent definitions. A term that is merely repeated with the same meaning is NOT an error --- only flag a term when its multiple definitions genuinely conflict. For example, if the Effective Date is defined in two places with conflicting dates in an agreement, the parties may disagree about which one controls. Consider a company hired to manage a property that is liable for any accidents occurring while the agreement is in effect. If the agreement defines the Effective Date as both March 1 and March 30, and an accident occurs on March 15, it becomes genuinely unclear who is responsible for that incident, the property owner or the management company. What should be a straightforward question can turn into a costly dispute. \\\\
\texttt{instruction:} List each occurrence separately along with each of the sections where the multiple definitions are located. \\\\
\texttt{schema:} \{"terms\_defined\_multiple\_times": [\{"term": "...", "location1": "...", "location2": "..."\}]\}

\end{tcolorbox}

\begin{tcolorbox}[colback=orange!5!white, colframe=orange!85!white,
                  title= {Incorrect Section, Article, or Paragraph References},
                  boxsep=2pt, left=4pt, right=4pt, top=2pt, bottom=2pt,
                  breakable]

\texttt{title:} Incorrect Section, Article, or Paragraph References\\\\

\texttt{definition:} Internal cross-references that assign an incorrect section, article, or paragraph of the agreement. Agreements are often reorganized during negotiation, with sections added, removed, or reordered. If an internal reference isn't updated to reflect those changes, it may end up pointing to the wrong section entirely. For example, suppose an agreement states that a party will face enhanced damages for breaches of ``Section 3,'' which covers confidentiality. During negotiation, the sections are reshuffled and Section 3 now covers product liability instead, but the reference is never updated. That party is now potentially exposed to enhanced damages for product liability incidents, a potentially broader and more expensive risk. \\\\
\texttt{instruction:} For each, list the incorrect reference used, the correct reference, and the section where the error is located. Use only the reference identifier (e.g. ``9.1''), without descriptive section names. List each reference as a SEPARATE item: if one cross-reference covers several sections, report each pair on its own (e.g. ``9.1 or 9.2 $\rightarrow$ 10.1 or 10.2'' becomes two items, 9.1 $\rightarrow$ 10.1 and 9.2 $\rightarrow$ 10.2), and if the same error recurs in multiple places, report each location separately. \\\\
\texttt{schema:} \{"incorrect\_section\_article\_paragraph\_references": [\{"wrong": "...", "correct": "...", "location": "..."\}]\}

\end{tcolorbox}

\begin{tcolorbox}[colback=orange!5!white, colframe=orange!85!white,
                  title= Incorrect Party References,
                  boxsep=2pt, left=4pt, right=4pt, top=2pt, bottom=2pt,
                  breakable]

\texttt{title:} Incorrect Party References\\\\
\texttt{definition:} Instances where a party is referred to by the wrong party name or role, such as ``Licensor'' instead of ``Licensee'' or ``Receiving Party'' instead of ``Disclosing Party''. Referencing the wrong party name in a contract can shift responsibilities in ways that a party may not want. If a clause requires a specific action, such as paying for shipping, but names the wrong party, the obligation could fall on that party regardless of what was originally discussed during contract negotiations. This type of error can end up costing a party time and money and potentially lead to a dispute over who is actually responsible. \\\\
\texttt{instruction:} List each item identifying the incorrect party reference, the correct party reference, and the section where the error is located. \\\\
\texttt{schema:} \{"incorrect\_party\_references": [\{"wrong": "...", "correct": "...", "location": "..."\}]\}

\end{tcolorbox}

\begin{tcolorbox}[colback=orange!5!white, colframe=orange!85!white,
                  title= Inconsistent Language,
                  boxsep=2pt, left=4pt, right=4pt, top=2pt, bottom=2pt,
                  breakable]

\texttt{title:} Inconsistent Language\\\\
\texttt{definition:} Language in the agreement that directly contradicts itself or other language elsewhere in the agreement.

When two provisions in a contract directly contradict each other, it creates uncertainty about which one actually applies. For example, if one section states that payment is due 30 days after receiving an invoice, but another states 45 days, neither party can be fully confident about when payment is expected. For the seller, this kind of ambiguity can complicate cash flow planning; they may be counting on payment at 30 days, while the buyer believes they have until 45. What starts as a drafting oversight can quickly become a source of friction or dispute. \\\\
\texttt{instruction:} For each occurrence, identify both sections where the conflicting language appears. \\\\
\texttt{schema:} \{"inconsistent\_terms": [\{"section1": "...", "section2": "..."\}]\}
\end{tcolorbox}
\end{document}